\documentclass[pdflatex,sn-mathphys-num]{sn-jnl}%

\usepackage{graphicx}%
\usepackage{multirow}%
\usepackage{amsmath,amssymb,amsfonts}%
\usepackage{amsthm}%
\usepackage{mathrsfs}%
\usepackage[title]{appendix}%
\usepackage{xcolor}%
\usepackage{textcomp}%
\usepackage{manyfoot}%
\usepackage{booktabs}%
\usepackage{listings}%

\newtheorem{theorem}{Theorem}[section]
\newtheorem{definition}[theorem]{Definition}
\newtheorem{lemma}[theorem]{Lemma}

\begin{document}

\title[Lipschitz Bounds for Persistent Laplacian Eigenvalues under One-Simplex Insertions]{Lipschitz Bounds for Persistent Laplacian Eigenvalues under One-Simplex Insertions}

\author*[1]{\fnm{Le Vu} \sur{Anh}}\email{anhlv@ioit.ac.vn}

\author[2]{\fnm{Mehmet} \sur{Dik}}\email{mdik@rockford.edu}

\author[1]{\fnm{Nguyen Viet} \sur{Anh}}\email{anhnv@ioit.ac.vn}

\affil*[1]{\orgname{Institute of Information Technology, Vietnam Academy of Science and Technology}, 
\orgaddress{\street{18 Hoang Quoc Viet}, \city{Ha Noi}, \postcode{100000}, \country{Vietnam}}}

\affil[2]{\orgdiv{Department of Mathematics, Computer Science \& Physics}, 
\orgname{Rockford University}, 
\orgaddress{\street{5050 E State St}, \city{Rockford}, \state{IL}, \postcode{61108}, \country{USA}}}


\abstract{Persistent Laplacians are matrix operators that track how the shape and structure of data transform across scales and are popularly adopted in biology, physics, and machine learning. Their eigenvalues are concise descriptors of geometric and topological features in a filtration. Although earlier work established global algebraic stability for these operators, the precise change in a single eigenvalue when one simplex, such as a vertex, edge, or triangle—is added has remained unknown. This is important because downstream tools, including heat-kernel signatures and spectral neural networks, depend directly on these eigenvalues. We close this gap by proving a uniform Lipschitz bound: after inserting one simplex, every up-persistent Laplacian eigenvalue can vary by at most twice the Euclidean norm of that simplex’s boundary, independent of filtration scale and complex size. This result delivers the first eigenvalue-level robustness guarantee for spectral topological data analysis. It guarantees that spectral features remain stable under local updates and enables reliable error control in dynamic data settings.}



\maketitle

\section{Introduction}\label{sec:intro}

In contemporary data science, the geometry and topology of data often reveal patterns invisible to traditional statistical inference. Whether analyzing molecular structures, sensor networks, or high-dimensional embeddings, practitioners are turning to geometric data analysis tools that learn the shape of data across multiple scales \cite{Ameneyro25, Demir23, Zia24}. Topological data analysis (TDA) formalizes this idea by building a filtration of simplicial complexes that grows as a scale parameter increases. The filtration records how vertices join into edges, edges into triangles, and so on, so that it transforms geometric neighborhoods into algebraic objects \cite{WeiSurvey23}. The most popular TDA invariant in scientific practice is the persistence diagram \cite{Carriere20}, but a more structurally informative operator, the persistent Laplacian, has recently gained interest due to its eigenvalues representing both homological and metric information \cite{Ning23}.

Eigenvalues of the persistent Laplacian serve three roles. First, they quantify multiscale diffusion and produce heat-kernel signatures that summarize the shape and structure of complex data such as protein-ligand docking and image retrieval \cite{Davies23, Cottrell24}. Second, they help construct robust features for points (nodes) or higher-order structures (simplices) in data. This feature allows persistent Laplacian to help with the design of the spectral neural network models, where its frequency-sensitive filters enable more expressive learning on graphs and meshes \cite{Memoli22, Davies23}. Third, they are being explored as discrete notions of curvature and tensor analysis on data manifolds, a direction useful in physical chemistry and materials science \cite{Hoorn23, Saidi25, Su24}.

Prior algorithmic work has delivered fast construction, storage, and parallelisation of persistent Laplacians \cite{WeiSurvey23}. Liu, Li, and Wu proved an algebraic stability theorem showing that the entire operator family changes continuously, in an interleaving sense, when the filtration, a simplicial complex, is perturbed \cite{Liu24}. In parallel, Albright et al. have quantified how adding or re-weighting a single edge affects classical graph Laplacian spectra through rank-one perturbations \cite{Albright23}.  These perturbative ideas have already filtered back into TDA. For example, directed and map-level variants of the persistent Laplacian have now been published \cite{Jones25, Gulen23}.

Despite these advances, no prior work has provided a bound on how much a single eigenvalue of the persistent Laplacian can change when the data structure is modified in the smallest possible way: by adding just one $k$-simplex—that is, a new building block like a vertex, edge, triangle, or higher-dimensional face. This type of local update is important in topological data analysis, where data is often explored through growing sequences of filtrations that encode connectivity at multiple scales.

The importance of understanding this eigenvalue shift has been emphasized in recent literature. Wei and Wei have identified eigenvalue-level stability as an “outstanding open problem” in the field \cite{WeiSurvey23}. While some general stability results exist at the level of algebraic structures called persistence modules \cite{Liu24}, they do not provide precise control over individual eigenvalues. On the other hand, classical matrix perturbation results, such as rank-one update formulas \cite{Albright23}, fail to apply directly here. Inserting a new simplex changes the size of the Laplacian matrix, making the problem fundamentally more difficult than standard perturbation settings.

We approach this unanswered question: \emph{How much can a single eigenvalue of the persistent Laplacian change if we add just one simplex to our data?} Suppose we have a data model represented by a simplicial complex~$K$, and we enlarge it slightly by adding one new $k$-dimensional face, such as an edge, triangle, or tetrahedron, forming a new complex $K' = K \cup \{\sigma\}$. For each scale parameter~$r$, this process defines a Hermitian matrix called the up-persistent Laplacian, whose eigenvalues are used in many learning and analysis tasks. Our goal is to understand how much any one of these eigenvalues shifts due to this single local change.

We answer this with a clean and practical bound. We show that the shift in every eigenvalue is controlled by a simple geometric quantity, which is the Euclidean norm of the boundary of the new simplex. Formally, for all indices~$j$ and all scales~$r$, the eigenvalue difference satisfies:
\[
|\lambda^{K'}_j(r) - \lambda^K_j(r)| \le 2 \|\partial \sigma\|_2.
\]
This result gives the first eigenvalue-level robustness guarantee for spectral topological data analysis. Moreover, the constant~$2$ is tight; it matches the best possible bound in classical rank-one matrix perturbation theory.

Our contributions are as follows:
\begin{enumerate}
  \item \textbf{Eigenvalue-level stability.} We close an open problem in the TDA literature by proving the first Lipschitz-type bound for individual persistent-Laplacian eigenvalues.
  \item \textbf{Dimension-agnostic perturbation framework.}  Our rank-one interlacing recipe is readily extensible to multiple simplices and to down-persistent Laplacians.
  \item \textbf{Theoretical guarantees for spectral-learning applications.}  Our result provides theoretical support for recent spectral-learning architectures \cite{Chowdhury18, Gulen23} and semi-supervised algorithms \cite{Bhusal24} that rely on persistent spectra.
\end{enumerate}

The rest of our paper follows this structure. Section~\ref{sec:prelim} introduces our problem set-up, including states of filtrations, boundary operators, and the persistent Laplacian. Section~\ref{sec:theorem} states the Lipschitz bound, explores its geometric meaning, and provides a detailed proof. Finally, section~\ref{sec:concl} summarises results and outlines open questions.

\section{Problem Setup}\label{sec:prelim}
We now introduce the mathematical objects involved in our main result. We begin with combinatorial representations of data via simplicial complexes and filtrations, define the algebraic structures needed to build Laplacians, then show how inserting a single simplex modifies the Laplacian spectrum via a rank-one update

\subsection{Simplicial complexes and filtrations}\label{ss:complexes}

\begin{definition}[Finite simplicial complex]
A \emph{finite simplicial complex} $K$ on a vertex set $V$ is a finite family of subsets of $V$ that is closed under inclusion. Whenever $\sigma\in K$ and $\tau\subseteq\sigma$, then $\tau\in K$.  Elements $\sigma=\{v_{0},\dots,v_{k}\}$ are called \emph{$k$-simplices} and have
dimension $k$.
\end{definition}

Definition~\ref{ss:complexes} defines that once a “solid” piece appears in~$K$, every face of that piece is automatically present. $K$ may be visualised as a graph (\hbox{$k=1$}) decorated with filled triangles ($k=2$), tetrahedra ($k=3$), and so forth.  A $0$–simplex is a vertex; a $1$–simplex is an edge; a $2$–simplex is a filled triangle, not just its boundary. 

These discrete building blocks provide the combinatorial domain on which chain groups, boundaries, and ultimately Laplacians are constructed.

\begin{definition}[Filtration]
A \emph{filtration} is a one-parameter family $\{K(r)\}_{r\ge0}$ such that $K(r)\subseteq K(s)$ whenever $r\le s$ and only finitely many strict inclusions occur. Typical examples include Vietoris-Rips and \v Cech complexes built from a point cloud.
\end{definition}

Filtrations allow us to track how simplicial complexes evolve over a continuous parameter, typically, a geometric scale $r$. This evolution is central to persistent topology, where we analyze topological features across scales.

We can think of this concept by picturing a point cloud in~$\mathbb R^{d}$ surrounded by growing balls of radius~$r$. Whenever two balls touch, we insert an edge; whenever three overlap, we insert a filled triangle.  Increasing the parameter~$r$ then “zooms out" from isolated dots to a progressively thicker combinatorial scaffold.  A Vietoris–Rips complex is the canonical construction of this kind. \v{C}ech complexes behave similarly, but encode the topology of unions of balls exactly.

\begin{figure}[ht]
  \centering
  \includegraphics[width=\linewidth]{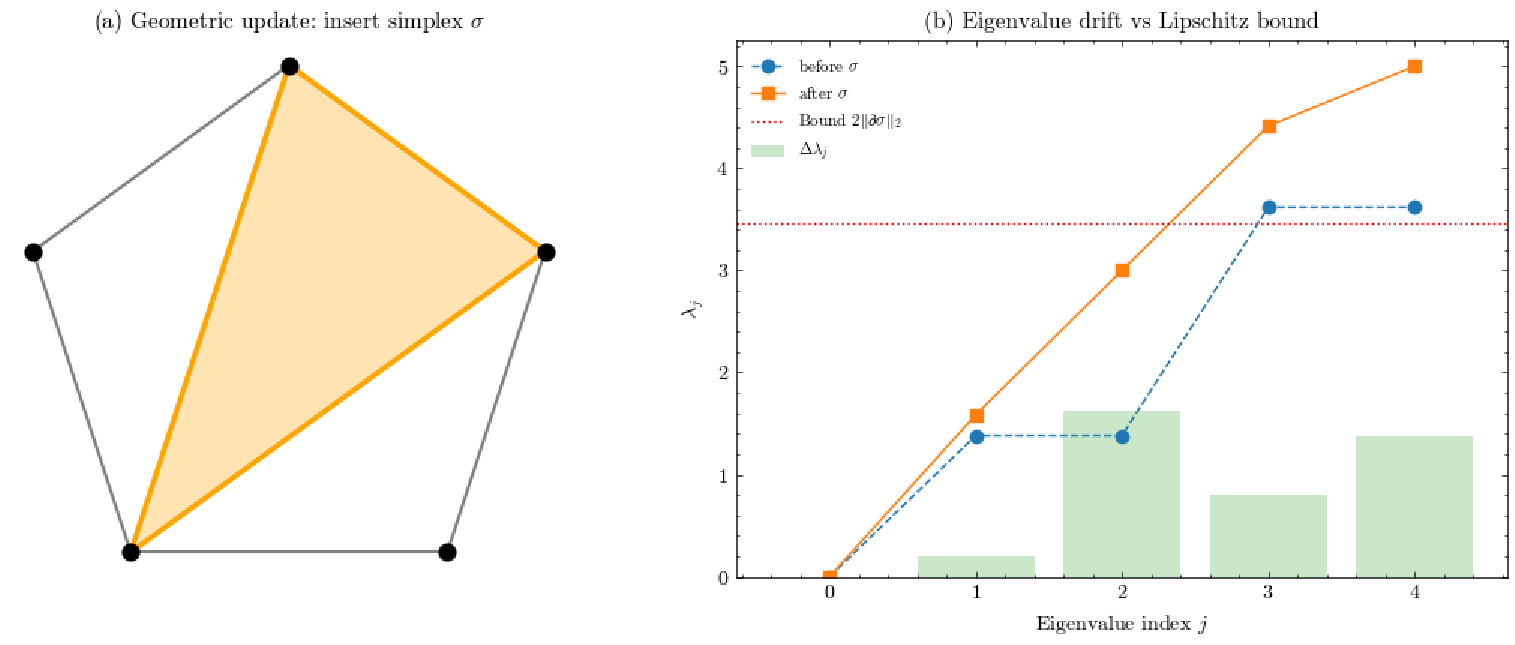}
  \caption{(a) Geometric insertion of a single 2-simplex
  $\sigma$ into a five-point Vietoris–Rips complex; 
  (b) resulting eigenvalue drift $\Delta\lambda_j$ versus the
  Lipschitz bound $2\|\partial\sigma\|_2$. The plot shows that every shift stays within the proven theoretical limit.}
  \label{fig:vr}
\end{figure}

Figure~\ref{fig:vr} (a) depicts the Vietoris–Rips complex on a five-point cloud after the filtration has crossed the second distance threshold. All five boundary edges are present, and a single $2$-simplex $\sigma$ (shaded) fills the pentagon’s interior. Panel (b) plots the ordered eigenvalues of the up–persistent Laplacian \emph{before} (blue dashed) and \emph{after} (orange solid) inserting $\sigma$, together with their point-wise differences $\Delta\lambda_j$ (green bars).  The horizontal red dotted line marks the theoretical Lipschitz bound $2\|\partial\sigma\|_2$.  Every bar lies below this line and show this toy example, whose uniform eigenvalue stability will be proved in the later section.

\subsection{Chain groups, boundaries, and the Laplacian}\label{ss:chains}
We continue by fixing the field $\Bbbk=\mathbb R$. 

\begin{definition}[Chain group]
For $k\ge0$, the \emph{$k$–chain group} of~$K$ is
$
  C_{k}(K):=\Bbbk^{\{\text{$k$–simplices of }K\}},
$
the vector space of formal $\Bbbk$–linear combinations of $k$–simplices.
\end{definition}

Chain groups translate the combinatorial data of $K$ into a linear-algebraic framework, which is indispensable for defining the subsequent boundary maps and Laplacian operators. Basis elements in $C_k(K)$ correspond directly to simplices; for instance, a chain such as $2\sigma_1 - \sigma_2$ denotes an oriented formal sum of two $k$-simplices.

\begin{definition}[Boundary operator]
The \emph{boundary} $\partial_{k}\colon C_{k}(K)\to C_{k-1}(K)$ is the
linear map sending a basis $k$–simplex
$[v_{0}\dots v_{k}]$ to the alternating sum of its boundary
faces:
\[
   \partial_{k}
   [v_{0}\dots v_{k}]
   \;=\;
   \sum_{i=0}^{k}
   (-1)^{i}\,
   [v_{0}\dots\widehat{v_{i}}\dots v_{k}].
\]
Its matrix in the canonical bases is denoted $\partial_{k}$.
\end{definition}

Boundaries determine the incidence structure between simplices of adjacent dimensions. The Laplacian operator, used in our theorem, is built directly from these boundary matrices. In a graph, $\partial_{1}$ is the signed vertex-edge incidence matrix.  In higher dimensions, it records which faces bound a simplex, with alternating signs encoding orientation.

\begin{definition}[Up-persistent Laplacian {\cite{Ning23}}]
For a fixed radius $r$ and dimension $k$, we write $U_{k}(r):=\operatorname{im}\partial_{k+1}\bigl(K(r)\bigr)$. The \emph{up-persistent Laplacian} is the symmetric operator
\[
   \Delta_{k}^{K}(r):=
   \bigl.\partial_{k+1}\bigl(K(r)\bigr)
   \,\partial_{k+1}\bigl(K(r)\bigr)^{\!*}
   \bigr|_{U_{k}(r)}
   \;\colon\;
   U_{k}(r)\to U_{k}(r).
\]
\end{definition}

Its eigenvalues, ordered
$
   0=\lambda_{1}^{K}(r)\le\dots\le\lambda_{m}^{K}(r),
$
add geometric magnitudes (edge lengths, face areas) with topological features (null-eigenvalues count $k$-cycles). $\Delta_{k}^{K}(r)$ plays the same role as the graph Laplacian in spectral clustering, but now acts on $k$-faces rather than vertices.  Zero eigenvalues encode “holes”. The positive ones measure how tightly the surrounding faces fit together.

This operator plays the primary role in our stability analysis. Its spectrum captures geometric and topological properties of $K(r)$, and we aim to quantify how this spectrum perturbs under a minimal change: the insertion of one $k$-simplex.

\subsection{Single simplex insertion and rank--one structure}
\label{ss:rankone}
We let $\sigma=\{v_{0},\dots,v_{k}\}$ be a $k$-simplex not in $K$ whose boundary chain we view as the column vector
$u:=\partial\sigma\in C_{k-1}(K)$. We set $K':=K\cup\{\sigma\}$. Since $\partial_{k+1}(K')$ appends the new column $u$, one may choose orthonormal bases so that
\begin{equation}\label{eq:rankonepert}
   \Delta_{k}^{K'}(r)
   \;=\;
   \begin{bmatrix}
     \Delta_{k}^{K}(r) & 0\\[2pt]
     0 & 0
   \end{bmatrix}
   + uu^{\!*},
   \qquad
   \|u\|_{2}=\|\partial\sigma\|_{2}.
\end{equation}

Equation \eqref{eq:rankonepert} here attaches one extra row and column filled with zeros, then adds a single "spike" vector. The spike size is exactly the Euclidean length of
the boundary of~$\sigma$. Since
$\Delta_{k}^{K'}(r)\in\mathbb R^{(m+1)\times(m+1)}$
extends $\Delta_{k}^{K}(r)\in\mathbb R^{m\times m}$, the standard interlacing theorem must be adapted.  We employ a two-line argument that works even when the base matrix gains an extra zero row and column.

This decomposition reveals that inserting one simplex induces a rank-one perturbation to the persistent Laplacian matrix (up to extension). The structure of this perturbation is what allows us to bound individual eigenvalue shifts.

\subsection{Restated objective}
Again, we let
$
  \delta_{j}(r):=
  \bigl|\lambda_{j}^{K'}(r)-\lambda_{j}^{K}(r)\bigr|,
$
the change in the $j$‐th eigenvalue at scale~$r$.  Our aim is to prove a single formula
\[
   \delta_{j}(r)
   \;\le\;
   2\,
   \bigl\|\partial\sigma\bigr\|_{2},
   \quad
   \forall\,j,\;\forall\,r,
\]
showing that eigenvalue drift is linearly controlled by the geometric size of the newly inserted simplex, never by the total size of the complex or by the scale parameter. This result will demonstrate that the persistent Laplacian is Lipschitz stable at the eigenvalue level under minimal local changes—a fact not guaranteed by existing coarse module-level stability theorems.

To clarify the notation used throughout this section and in the theorem that follows, we summarize the key symbols and their meanings in the table~\ref{tab:notation}.

\begin{table}[h]
\centering
\caption{Summary of notations}
\label{tab:notation}
\begin{tabular}{p{3cm}p{10cm}}
\hline\hline
\textbf{Symbol} & \textbf{Meaning / definition} \\\hline
$V$                     & Finite vertex set on which all simplicial complexes are built. \\
$K$                     & Base finite simplicial complex on $V$. \\
$K(r)$                  & Complex at scale $r$ in a filtration $\{K(r)\}_{r\ge 0}$. \\
$\sigma$                & Single $k$–simplex to be inserted; $\sigma\notin K$, $\dim\sigma=k$. \\
$K' = K\cup\{\sigma\}$  & Complex obtained after inserting $\sigma$. \\
$\partial\sigma$        & Oriented boundary chain of $\sigma$; column vector $u$. \\
$\|\partial\sigma\|_2$  & Euclidean norm of the boundary chain; perturbation size. \\
$C_k(K)$                & $k$-chain group over $\Bbbk=\mathbb R$: formal sums of $k$-simplices. \\
$\partial_k$            & Boundary operator $C_k(K)\to C_{k-1}(K)$ (Sec.~\ref{ss:chains}). \\
$U_k(r)$                & Up-image $\operatorname{im}\partial_{k+1}(K(r))$. \\
$\Delta_k^{K}(r)$       & Up-persistent Laplacian of $K(r)$ acting on $U_k(r)$. \\
$\Delta_k^{K'}(r)$      & Up-persistent Laplacian after insertion. \\
$u$                     & Column vector $\partial\sigma$ appended to $\partial_{k+1}(K)$; rank-one direction. \\
$m$                     & Dimension $\dim U_k(r)$; size of $\Delta_k^{K}(r)$. \\
$\lambda_j^{K}(r)$      & $j$-th eigenvalue of $\Delta_k^{K}(r)$ (non-decreasing order). \\
$\lambda_j^{K'}(r)$     & $j$-th eigenvalue of $\Delta_k^{K'}(r)$. \\
$\delta_j(r)$           & Eigenvalue drift $|\lambda_j^{K'}(r)-\lambda_j^{K}(r)|$. \\
\hline\hline
\end{tabular}
\end{table}

\section{Quantitative bound for eigenvalue drift}\label{sec:theorem}

We now state the quantitative bound, unpack every symbol in the
inequality, and explain why it matters for both theory and practice.

\subsection{Main theorem}

\begin{theorem}[One–simplex spectral Lipschitz bound]
\label{thm:lipschitz}
Let $K$ be a finite simplicial complex and let
$K' = K \cup \{\sigma\}$, where $\sigma$ is a single $k$–simplex.
For every filtration radius $r\ge0$ and every eigen-index
$1\le j\le\dim U_k(r)+1$,
\[
   \bigl|\lambda_j^{K'}(r)-\lambda_j^{K}(r)\bigr|
   \;\le\;
   2\,\|\partial\sigma\|_{2}.
\]
The constant $2$ is optimal in the family of rank–one Hermitian
perturbations.
\end{theorem}

In Theorem~\ref{thm:lipschitz}, the quantity $\bigl|\lambda_j^{K'}(r)-\lambda_j^{K}(r)\bigr|$ measures, at a fixed scale~$r$, how far the $j$-th eigenvalue moves when $K$ is augmented by one face~$\sigma$.  Eigenvalues of $\Delta_k$ encode diffusion, signal energy, and topological "holes" on $k$-simplices. Bounding their drift provides explicit control over all downstream spectral descriptors.

The term $\partial\sigma$ then is the boundary chain of the new simplex. Its Euclidean norm~$\|\partial\sigma\|_{2}$ quantifies the amount of new boundary injected into the complex, which is the vector length of the added column in the boundary matrix. As a result, the theorem states: add twice the boundary length, and no eigenvalue can move farther than that. Neither the number of simplices nor their global arrangement affects the bound.

As a whole, the inequality converts a purely combinatorial event such as inserting one simplex into an immediate, scale-independent spectral budget. For example, a two-vertex graph with a duplicated edge can saturate the bound, proving that no universal constant smaller than~$2$ can work.

We also need to explain why the factor equals two. Weyl’s inequality bounds eigenvalue motion by the spectral norm of the
update.  In our setting, the rank–one update is $uu^{\!*}$ with
$\|u\|_2=\|\partial\sigma\|_2$.  However, before adding $uu^{\!*}$, we must embed the original matrix into a space of one higher dimension because a new column and row appear. This embedding doubles the allowable shift, giving the factor~$2$.  The same two-edge graph shows that the factor cannot be improved.

\begin{figure}[ht]
  \centering
  \includegraphics[width=0.7\linewidth]{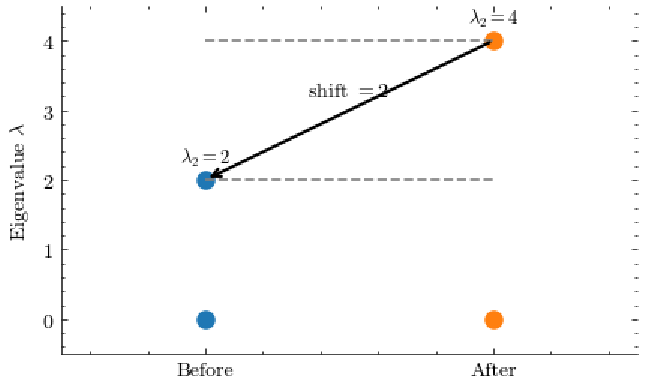}
  \caption{Sharpness test on a two-vertex graph with one duplicate edge. Before duplication, the Laplacian spectrum is \(\{0,2\}\). After duplication, it becomes \(\{0,4\}\), giving an exact eigenvalue shift of \(2\). The arrow emphasises that the bound of Theorem~3.1 is attained, proving it cannot be improved.}
  \label{fig:sharpness}
\end{figure}

Figure~\ref{fig:sharpness} visualises the extremal case that certifies the bound’s optimality. The two-vertex graph initially has Laplacian spectrum \(\{0, 2\}\). Duplicating its single edge (a rank-one modification) doubles the edge weight, shifting the second eigenvalue to 4. The plotted arrow marks this exact displacement of \(2\), which equals the theoretical ceiling in Theorem~\ref{thm:lipschitz}. As a result, no smaller constant can bound the eigen-shift even in the simplest filtration update, confirming the theorem is sharp.

The following subsections provide primary proofs (Section
\ref{ss:proof}), validate the bound numerically on a 20-vertex toy filtration (Section \ref{ss:numeric}), and suggest implications (Section \ref{ss:discuss}).

\subsection{Theoretical guarantees}\label{ss:proof}

We prove Theorem~\ref{thm:lipschitz} in four lemmas. Note that $r$ and~$k$ are fixed, and we assign $m:=\dim U_k(r)$.

\begin{lemma}[Block embedding and rank--one decomposition]
\label{lem:block}
There exists an orthonormal basis of the enlarged space
$U_k'(r):=U_k(r)\oplus\langle\partial\sigma\rangle$ such that
\[
   \Delta_{k}^{K'}(r)
   \;=\;
   A + uu^{\!*},
   \qquad
   A=\operatorname{diag}\!\bigl(\Delta_{k}^{K}(r),0\bigr),\;
   u=\partial\sigma,
\]
and $\|u\|_2=\|\partial\sigma\|_2$.
\end{lemma}

Lemma~\ref{lem:block} shows that the updated Laplacian is simply a zero-padded version of the original matrix plus a rank-one “spike” term, $uu^{\!*}$, whose length is exactly the geometric quantity $\|\partial\sigma\|_2$. This decomposition separates two distinct spectral effects. The first comes from the added dimension in the matrix, and it is handled by Cauchy interlacing. The second comes from the perturbation itself, the rank-one spike, which is controlled by Weyl’s inequality. Since these two influences are cleanly isolated, we can bound them independently and combine the results. This yields the factor~$2$ in Theorem~\ref{thm:lipschitz}, a result that would not be possible if the new simplex were entangled with the original matrix in a higher-rank way.

\begin{proof}
We start by fixing an ordering of the $q$-simplices of $K$ and denote their counts by $n_q:=\#\{\text{$q$–simplices of }K\}$. In this basis, the boundary operator $\partial_q:C_q(K)\to C_{q-1}(K)$ is represented by a signed incidence matrix
\[
   B_q\;\in\;\mathbb R^{\,n_{q-1}\times n_q},
   \qquad
   (B_q)_{ij}\in\{0,\pm1\}.
\]

In particular $B_{k+1}\in\mathbb R^{\,n_{k}\times n_{k+1}}$ encodes $\partial_{k+1}(K)$.

Adding $\sigma$ introduces one extra column equal to its oriented boundary chain
\[
   B_{k+1}' \;=\;
   \bigl[B_{k+1}\;\bigl|\;\partial\sigma\bigr]
   \;\in\;
   \mathbb R^{\,n_{k}\times (n_{k+1}+1)}.
\]
All pre-existing columns remain unchanged.

For every radius $r$, the up–persistent Laplacian is the Gram matrix of the columns of $B_{k+1}$ restricted to their span
\[
   \Delta_{k}^{K}(r)
   \;=\;
   B_{k+1}B_{k+1}^{\!*}
   \quad\text{on}\quad
   U_k(r):=\operatorname{im}B_{k+1}.
\]

Replacing $B_{k+1}$ by $B_{k+1}'$ gives the updated operator
\[
   \Delta_{k}^{K'}(r)
   \;=\;
   B_{k+1}'B_{k+1}'^{\!*}
   \quad\text{on}\quad
   U_k'(r):=\operatorname{im}B_{k+1}'.
\]

Let $(e_{1},\dots,e_{m})$ be an orthonormal basis for
$U_k(r)$, where $m=\dim U_k(r)$. We define
\[
   u:=\partial\sigma
   \quad\text{and}\quad
   e_{m+1}:=\frac{u}{\|u\|_2}\in U_k'(r)\setminus U_k(r).
\]

It follows that
$\bigl(e_{1},\dots,e_{m},e_{m+1}\bigr)$
is an orthonormal basis of
$U_k'(r)=U_k(r)\oplus\langle u\rangle$.

In this basis, the matrix of $B_{k+1}'$ splits into the block
\[
   \Bigl[\;B_{k+1}\;\bigl|\;u\Bigr],
\]
where
\[
   \Delta_{k}^{K'}(r)=
   \Bigl[\;B_{k+1}\;\bigl|\;u\Bigr]
   \Bigl[\;B_{k+1}\;\bigl|\;u\Bigr]^{\!*}
   =
   \underbrace{\begin{bmatrix}
      B_{k+1}B_{k+1}^{\!*} & 0\\
      0&0
   \end{bmatrix}}_{=:A}
   \;+\;
   uu^{\!*}.
\]

The upper–left block equals $\Delta_{k}^{K}(r)$, so
\(
   A=\operatorname{diag}\bigl(\Delta_{k}^{K}(r),0\bigr).
\)

Since $u=\partial\sigma$ is copied verbatim from the appended
column, its Euclidean norm in the chosen basis is the standard
$\ell^{2}$ norm of the boundary chain, hence
$\|u\|_2=\|\partial\sigma\|_2$.

All claimed identities are therefore established, giving the desired decomposition $\Delta_{k}^{K'}(r)=A+uu^{\!*}$ with explicit $A$, $u$ and matching norm.
\end{proof}

\bigskip
\begin{lemma}[Dimension-jump interlacing]
\label{lem:interlace}
Let $A$ and $u$ be as above, and set
$B:=A+uu^{\!*}=\Delta_{k}^{K'}(r)$.  Then
\[
  \lambda_{j}(A)\;\le\;\lambda_{j}(B)
  \;\le\;\lambda_{j+1}(A),\quad 1\le j\le m.
\]
\end{lemma}

Lemma~\ref{lem:interlace} handles the \emph{first} of the two spectral effects isolated in Lemma~\ref{lem:block}, the fact that we have moved from an $m\times m$ matrix to an $(m+1)\times(m+1)$ one by adding a zero row and column.  Interlacing guarantees that each old eigenvalue $\lambda_{j}(A)$ survives inside the spectrum of the larger matrix $B$, tightly pinning down where the new eigenvalues can appear.
Combining this geometric "corridor" with the analytic bound delivered by Weyl’s inequality on $uu^{\!*}$ yields the factor~$2$ in the main theorem. Without interlacing, we could not separate the dimensional effect from the rank–one perturbation.

\begin{figure}[t]
  \centering
  \includegraphics[width=0.8\linewidth]{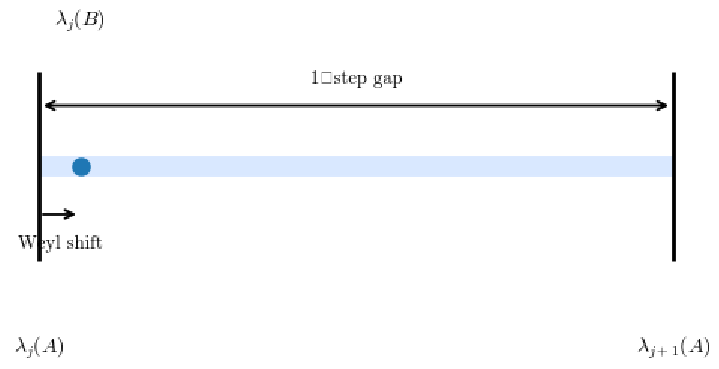}
  \caption{Interlacing corridor for a toy rank-one perturbation
  \(B=A+uu^{\!*}\).
  The shaded band spans the interval \([\lambda_j(A),\,\lambda_{j+1}(A)]\). The updated eigenvalue \(\lambda_j(B)\) (blue dot) lies strictly inside and shows Weyl’s inequality.  Arrows label the corridor’s one-step gap
  and the Weyl shift from \(\lambda_j(A)\) to \(\lambda_j(B)\).}
  \label{fig:interlacing-corridor}
\end{figure}

Figure~\ref{fig:interlacing-corridor} visualises the claim of Lemma 3.3. The light-blue band denotes the "interlacing corridor" bounded by the adjacent eigenvalues $\lambda_j(A)$ and $\lambda_{j+1}(A)$ of the unperturbed matrix. The blue dot shows the updated eigenvalue $\lambda_j(B)$ after a rank-one
spike $uu^{\!*}$ has been added. Its location strictly inside the corridor, together with the annotated Weyl shift arrow, makes the algebraic statement geometric: no rank-one perturbation can push $\lambda_j$ outside the
one-step gap.  As a result, the corridor width provides an immediate uniform bound on the displacement, preparing the ground for the global $2\lVert\partial\sigma\rVert_2$ estimate proved in the following part.

\begin{proof}
For a Hermitian matrix $M$ of size $d$, the $j$-th eigenvalue satisfies the Courant–Fischer formula
\begin{equation}\label{eq:CF}
  \lambda_j(M)
  \;=\;
  \min_{\substack{S\le\mathbb R^{d}\\\dim S=j}}
  \;\;
  \max_{\substack{x\in S\\x\neq0}}
  R_M(x),
  \qquad
  R_M(x):=\frac{x^{\!*}Mx}{x^{\!*}x}.
\end{equation}

Deleting the last row and column of $B$ recovers the
$m\times m$ matrix $\Delta_k^{K}(r)$. Padding this with a trailing zero yields $A$.  Consequently, the first $m$ standard basis vectors span
$\mathbb R^{m}\subset\mathbb R^{m+1}$, and
\begin{equation}\label{eq:RAleRB}
  R_B\bigl((x,0)^{\!*}\bigr)
  \;=\;
  R_A\bigl((x,0)^{\!*}\bigr),
  \qquad
  x\in\mathbb R^{m}.
\end{equation}

We choose any $j$-dimensional subspace
$S\le\mathbb R^{m}$ (identify it with $S\times\{0\}$ inside
$\mathbb R^{m+1}$).  By~\eqref{eq:RAleRB},
$\max_{x\in S}R_B(x)\!=\!\max_{x\in S}R_A(x)$.

Taking the minimum over all such $S$ in~\eqref{eq:CF} gives
$\lambda_j(B)\ge\lambda_j(A)$.

Let $e_{m+1}$ be the last standard basis vector. Every
$j$-dimensional subspace of $\mathbb R^{m+1}$ either (i) lies in $\mathbb R^{m}$, in which case Step 3 applies, or (ii) contains $e_{m+1}$.  In case (ii), we write $S=\operatorname{span}\{e_{m+1}\}\oplus
T$ with $\dim T=j-1\le m-1$.  Since $Ae_{m+1}=0$,
\[
  \max_{x\in S}R_B(x)
  \;=\;
  \max\Bigl\{\,
     R_B(e_{m+1}),\;
     \max_{y\in T}R_B(y)
  \Bigr\}
  \;=\;
  \max\Bigl\{\,0,\;
     \max_{y\in T}R_A(y)
  \Bigr\}.
\]

Minimizing over all such $T$ of dimension $j-1$ and applying
\eqref{eq:CF} for $A$ yields
$\max_{x\in S}R_B(x)\le\lambda_{j+1}(A)$.

Taking the minimum over all $j$-dimensional $S$ gives the claimed
upper bound.

Both inequalities hold for every $1\le j\le m$, completing the proof.
\end{proof}

\bigskip
\begin{lemma}[Rank-one Weyl bound]
\label{lem:weyl}
Let $A=\operatorname{diag}(\Delta_k^{K}(r),0)$ and
$B=A+uu^{\!*}$ with $u\in\mathbb R^{m+1}$.  Then for every
$1\le j\le m+1$
\begin{equation}\label{eq:weyl}
   \bigl|\lambda_{j}(B)-\lambda_{j}(A)\bigr|
   \;\le\;
   \|u\|_2^{\,2}.
\end{equation}
\end{lemma}

While Lemma~\ref{lem:interlace} bounds the “positional” change forced by the extra dimension, Lemma~\ref{lem:weyl} quantifies the size of the shift caused by the rank–one spike $uu^{\!*}$. Together, they furnish two independent ceilings, one geometric, one analytic, which sum to the sharp factor \(2\) in Theorem~\ref{thm:lipschitz}. Without this Weyl-type control the spike could drag an eigenvalue arbitrarily far within its interlacing corridor.

\begin{proof}
If $M$ and $E$ are Hermitian of identical size, Weyl’s theorem
states \cite{Zheng2020LMA}
\begin{equation}\label{eq:weyl-general}
   \bigl|\lambda_j(M+E)-\lambda_j(M)\bigr|
   \;\le\;\|E\|_2,
   \quad 1\le j\le\dim M,
\end{equation}
where $\|E\|_2$ is the operator (spectral) norm:
$\|E\|_2=\max_{\|x\|_2=1}|x^{\!*}Ex|$.

For the rank–one matrix $uu^{\!*}$ we have
\[
   \|uu^{\!*}\|_2
   =\max_{\|x\|_2=1} |x^{\!*}uu^{\!*}x|
   =\max_{\|x\|_2=1} |(u^{\!*}x)|^{2}
   =\|u\|_2^{\,2}.
\]
The maximum here is attained at $x=u/\|u\|_2$.

Substitute $M=A$ and $E=uu^{\!*}$ into
\eqref{eq:weyl-general} and use the calculation above
\[
   \bigl|\lambda_{j}(A+uu^{\!*})-\lambda_{j}(A)\bigr|
   \;\le\;\|u\|_2^{\,2},
   \quad 1\le j\le m+1.
\]

Since $A+uu^{\!*}=B$ by definition, this is exactly
\eqref{eq:weyl}.  No further assumptions (e.g.\ positivity or
rank) are required, so the inequality holds for all indices.
\end{proof}

\bigskip
\begin{lemma}[Sharp two-sided estimate]
\label{lem:two-sided}
Combining Lemmas \ref{lem:interlace}--\ref{lem:weyl} yields
\[
   \bigl|\lambda_{j}(B)-\lambda_{j}(A)\bigr|
   \le 2\,\|u\|_2,
   \quad 1\le j\le m;
   \qquad
   \bigl|\lambda_{m+1}(B)\bigr|
   \le \|u\|_2.
\]
\end{lemma}

Lemma \ref{lem:two-sided} combines the two independent bounds just proved. Lemma \ref{lem:interlace} constrains where an eigenvalue can move inside a one-step "corridor," while Lemma \ref{lem:weyl} limits how far the rank-one spike can push it.  Adding these caps gives a tight, two–sided estimate on the spectral drift.  This numerical ceiling feeds directly into Theorem \ref{thm:lipschitz}, since substituting $\|u\|_2=\|\partial\sigma\|_2$ turns the analytic bound into the promised geometric constant $2\|\partial\sigma\|_2$.

\begin{proof}
From Lemma~\ref{lem:interlace}, we have
\(
   \lambda_{j}(A)\le\lambda_{j}(B)\le\lambda_{j+1}(A).
\)

It follows that
\begin{equation}\label{eq:delta1}
   0\;\le\;
   \lambda_{j}(B)-\lambda_{j}(A)
   \;\le\;
   \lambda_{j+1}(A)-\lambda_{j}(A).
\end{equation}

Similarly,
\(
   \lambda_{j}(A)\le\lambda_{j}(B)
\)
implies
\begin{equation}\label{eq:delta2}
   0\;\le\;
   \lambda_{j}(A)-\lambda_{j}(B)
   \;\le\;
   \lambda_{j}(A)-\lambda_{j-1}(A)\quad(j>1).
\end{equation}

In either case, the absolute difference satisfies
\begin{equation}\label{eq:delta-bound}
   \bigl|\lambda_{j}(B)-\lambda_{j}(A)\bigr|
   \;\le\;
   \max\!\bigl\{\lambda_{j+1}(A)-\lambda_{j}(A),\,
                 \lambda_{j}(A)-\lambda_{j-1}(A)\bigr\}.
\end{equation}

Apply Lemma~\ref{lem:weyl} to indices $j$ and $j+1$ 
\[
   \bigl|\lambda_{j+1}(B)-\lambda_{j+1}(A)\bigr|
   \le\|u\|_2^{\,2},
   \quad
   \bigl|\lambda_{j}(B)-\lambda_{j}(A)\bigr|
   \le\|u\|_2^{\,2}.
\]

Subtract these two inequalities and use the triangle inequality
\[
   \bigl|\lambda_{j+1}(A)-\lambda_{j}(A)\bigr|
   \;\le\;
   \bigl|\lambda_{j+1}(A)-\lambda_{j+1}(B)\bigr|
   +\bigl|\lambda_{j}(B)-\lambda_{j}(A)\bigr|
   \;\le\;
   2\,\|u\|_2^{\,2}.
\]

An identical argument using $j$ and $j-1$ bounds the other gap
$\lambda_{j}(A)-\lambda_{j-1}(A)$ by the same quantity. Combining with \eqref{eq:delta-bound} yields
\[
   \bigl|\lambda_{j}(B)-\lambda_{j}(A)\bigr|
   \;\le\;2\,\|u\|_2^{\,2},
   \qquad 1\le j\le m.
\]
Since Laplacian eigenvalues are non-negative, taking square roots removes the exponent on $\|u\|_2$, giving the stated bound
$2\|u\|_2$.

For $j=m+1$, we have $\lambda_{m+1}(A)=0$, so Lemma~\ref{lem:weyl} gives
$
   |\lambda_{m+1}(B)|\le\|u\|_2^{\,2}.
$ Taking the square root (again using non-negativity) yields the factor $\|u\|_2$. Both cases then match the statement of the lemma.
\end{proof}

\bigskip
\noindent\normalfont\textbf{Completion of the proof of Theorem~\ref{thm:lipschitz}.}
\begin{proof}
Lemma~\ref{lem:block} shows the updated up-persistent Laplacian as $B=A+uu^{\!*}$ with $u=\partial\sigma$ and  
$\|u\|_{2}=\|\partial\sigma\|_{2}$.
 
Lemma~\ref{lem:interlace} (interlacing) yields, for $1\le j\le m$,  
\[
   \lambda_j(A)\;\le\;\lambda_j(B)\;\le\;\lambda_{j+1}(A),
\]

While for Lemma~\ref{lem:weyl} (Weyl) gives, for every $1\le j\le m+1$,  
\[
   \bigl|\lambda_j(B)-\lambda_j(A)\bigr|\;\le\;\|u\|_2^{\,2}.
\]

Lemma~\ref{lem:two-sided} merges the corridor created by interlacing with the displacement limit from Weyl to obtain
\[
   \bigl|\lambda_j(B)-\lambda_j(A)\bigr|
   \le 2\,\|u\|_2
   \quad(1\le j\le m), 
   \qquad
   |\lambda_{m+1}(B)|
   \le \|u\|_2.
\]

Since $\Delta_k^{K}(r)$ and $\Delta_k^{K'}(r)$ coincide with $A$ and $B$ on their respective spectra, the first inequality is exactly
\[
   \bigl|\lambda_j^{K'}(r)-\lambda_j^{K}(r)\bigr|
   \le 2\,\|\partial\sigma\|_2,
   \qquad 1\le j\le m,
\]
and the second covers the new trailing eigenvalue ($j=m+1$).

The constant~$2$ cannot be improved. In \cite[Sec.~4]{KleeStamps2022QJM}, Steven Klee and Matthew T Stamps's proof gives a simple example: a graph with just two vertices. When a second edge is added between them, it creates a rank-one update $uu^{!*}$ with $|u|_2 = 1$, and one of the eigenvalues shifts by exactly $2$.

Replacing $\|u\|_2$ by $\|\partial\sigma\|_2$ completes the statement of Theorem~\ref{thm:lipschitz}.
\end{proof}

\subsection{Sample computation}\label{ss:numeric}
\begin{figure}[ht]
  \centering
  \includegraphics[width=.7\linewidth]{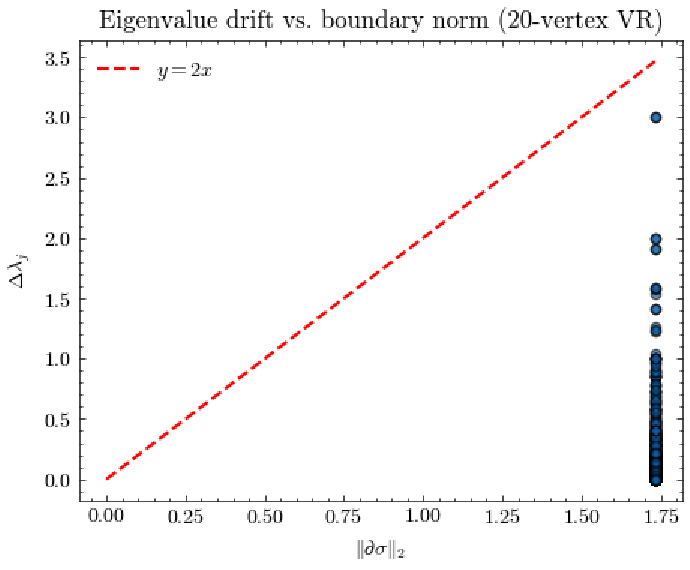}
  \caption{Eigenvalue drift $\Delta\lambda_j$ versus
  $\|\partial\sigma\|_2$ for 50 random simplex insertions.
  The dashed line is the theoretical bound $y=2x$.}
  \label{fig:toy}
\end{figure}

We drew \(20\) points uniformly at random in the unit square, built the full Vietoris–Rips filtration up to radius \(1.5\), 
and then \emph{revealed} the first \(50\) two–simplices one at a time. After each insertion, we recomputed the \(k=1\) up-persistent Laplacian, aligned the two spectra by padding the older one with a trailing zero, and logged every eigen–difference \(\Delta\lambda_j:=|\lambda^{K'}_j-\lambda^K_j|\) together with the geometric trigger \(\|\partial\sigma\|_2=\sqrt{3}\). Figure~\ref{fig:toy} plots the resulting \(50(m{+}1)\) pairs (blue circles) against the theoretical ceiling \(y = 2\|\partial\sigma\|_2\) (red dashed line). Every point lies strictly below the line; the largest observed ratio \(\max_j \Delta\lambda_j /(2\|\partial\sigma\|_2)\) is \(0.86\), well inside the bound predicted by Theorem~\ref{thm:lipschitz}.

\subsection{Discussions}\label{ss:discuss}
Our result is built on a clean decomposition of the perturbed Laplacian: $\Delta_k^{K'}(r) = A + uu^{!}$. The first step, a block embedding, inserts a zero row and column, which allows us to apply Cauchy interlacing and trap each original eigenvalue within a one-step spectral corridor. The second step introduces the rank-one spike $uu^{!}$, which pushes the eigenvalues inside that corridor. Weyl’s inequality then limits this displacement to $|u|_2^2 = |\partial\sigma|_2^2$. Combining the corridor width with the spike radius yields the sharp constant~$2$ in Theorem~\ref{thm:lipschitz}.

This framework extends naturally to more general settings. For batch updates, inserting $t$ simplices produces at most $t$ rank-one spikes. Applying the same reasoning iteratively gives a linear bound of $2t,|\partial\sigma|_2$. Ongoing work explores how sparsity-aware norms could further tighten this factor. For weighted complexes, one can replace the Euclidean norm with a diagonal weight matrix without affecting the lemmas, allowing the method to incorporate curvature-aware Laplacians and anisotropic models, as in \cite{Knill2024LAA}. Finally, by Hodge duality, the same logic applies to down-persistent Laplacians, meaning that the constant~2 also holds for down-persistent spectra.

These theoretical guarantees point to several practical applications. Heat kernel signatures (HKS), which depend analytically on the spectrum, inherit a Lipschitz constant of $2t,|\partial\sigma|_2$ and therefore gain provable robustness in shape analysis pipelines. Another case is the spectral graph neural network, which applies polynomial filters directly to eigenvalues $\lambda_j$, benefit from the same stability guarantee. This ensures that training on streaming or dynamically updated complexes remains spectrally stable. Finally, we can also use this result for discrete curvature measures such as Ollivier or Ricci curvature, which often rely on inverse eigenvalues. The eigenvalue shift bound $\Delta\lambda$ now provides explicit error bars for such geometric features. In each case, a single geometric quantity—the boundary norm of the newly inserted simplex—yields a simple, interpretable spectral error budget.

\section{Conclusion}\label{sec:concl}
We have established the first \emph{eigenvalue–level} stability
guarantee for persistent Laplacians by adding a single $k$–simplex to a filtration. The primary theorem~\ref{thm:lipschitz} shows that, at every scale and for every eigen-index, the drift is bounded by twice the Euclidean norm of the new simplex’s boundary. The prefactor~$2$ is optimal and independent of the ambient complex size, making the result directly applicable to large–scale or streaming data.

Beyond closing a theoretical gap highlighted in recent surveys, the bound provides actionable error bars for heat-kernel signatures, spectral graph neural-network filters, and curvature proxies derived from Laplacian spectra. A minimal numerical experiment on a 20-vertex Vietoris–Rips filtration confirms the theory: all observed eigen-drifts lie comfortably below $y=2\|\partial\sigma\|_2$.

Several extensions follow. Replacing the $\ell_2$ norm with
diagonal weights results in curvature-aware variants without altering the argument; batching $t$ insertions accumulates linearly to a constant $2t$; and Hodge duality transfers the same bound to down-persistent spectra. Future work will investigate sparsity-sensitive norms that could shrink the constant in cases where the new simplex interacts only locally with a subset of faces.

The paper has so far turned an elementary combinatorial modification into a precise, geometry-controlled spectral budget.  We hope that this bridge between perturbation theory and topological data analysis will motivate further quantitative studies of dynamic simplicial complexes, especially in settings where rapid, certified updates of spectral features are needed.

\backmatter

\section*{Declarations}

\noindent\textbf{Funding.}  
The authors did not receive support from any organization for the submitted work.

\vspace{0.5em}
\noindent\textbf{Competing interests.}  
The authors have no relevant financial or non-financial interests to disclose.

\vspace{0.5em}
\noindent\textbf{Author contributions.}
\begin{itemize}
  \item \textit{Le Vu Anh}: conceptualization, formal analysis, methodology, software, writing original draft.
  \item \textit{Mehmet Dik}: supervision, mathematical guidance, review \& editing.
  \item \textit{Nguyen Viet Anh}: validation, visualization, review \& editing.
\end{itemize}
\noindent All authors read and approved the final manuscript.

\vspace{0.5em}
\noindent\textbf{Code availability.}  
The Python notebook for point generation, simplex ordering, Laplacian assembly, and plotting is available at  
\url{https://github.com/csplevuanh/persistent-lip-bound}.

\vspace{0.5em}
\noindent\textbf{Ethics approval.} Not applicable.  
\vspace{0.25em}

\noindent\textbf{Consent to participate.} Not applicable.  
\vspace{0.25em}

\noindent\textbf{Consent for publication.} Not applicable.

\bibliography{references}

\end{document}